# Information Loss in LLMs' Multilingual Translation: The Role of Training Data, Language Proximity, and Language Family

Yumeng Lin[1], Xufeng Duan[1], David Haslett[2], Yige Chen[1], Zhenguang G. Cai[1,3]

[1] Department of Linguistics and Modern Languages, The Chinese University of Hong Kong, Hong Kong, Hong Kong SAR, China.

[2] Division of Social Science, The Hong Kong University of Science and Technology, Hong Kong SAR, China

[3] Brain and Mind Institute, The Chinese University of Hong Kong, Hong Kong, Hong Kong SAR, China.

Correspondence should be addressed to Zhenguang G. Cai, Department of Linguistics and Modern Languages, Leung Kau Kui Building, The Chinese University of Hong Kong, Shatin, Hong Kong SAR; zhenguangcai@cuhk.edu.hk.

**Abstract:** Large language models have achieved impressive progress in multilingual translation, yet they continue to face challenges with certain language pairs—particularly those with limited training data or significant linguistic divergence from English. This study systematically investigates how training data, language proximity, and language family affect information loss in multilingual translation. We evaluate two large language models, GPT-4 and Llama 2, by performing round-trip translations. Translation quality was assessed using BLEU scores and BERT similarity metrics. Our results reveal a robust interaction between training data size and language distance: while abundant training data can mitigate the effects of linguistic divergence, languages structurally closer to English consistently yield higher translation quality in low-resource conditions. Among various distance metrics, orthographic, phylogenetic, syntactic, and geographical distances emerge as strong predictors of translation performance. Language family also exerts an independent influence. These findings contribute to a deeper understanding of the linguistic constraints shaping multilingual translation in large language models, emphasizing that translation quality is shaped not only by data volume but also by structural and typological relationships between languages.

## 1 INTRODUCTION

Large Language Models (LLMs) demonstrated advanced multilingual capabilities. Among these, machine translation stands out as a particularly important application, enabling cross-linguistic communication and fostering global accessibility to information. However, multilingual machine translation still faces challenges, as it requires models not only to translate across diverse languages but also to maintain semantic alignment between them (Team et al., 2022). A growing body of research has evaluated LLM performance in multilingual translation, revealing that state-of-the-art models exhibit strong proficiency in high-resource languages (e.g., Bang et al., 2023; Jiao et al., 2023). For instance, the 7B Llama 2 model achieves BLEU scores above 10 when translating into all languages encountered during training (Diandaru et al., 2024). Similarly, GPT-4 has been shown to perform comparably to junior human translators for high-resource language pairs (Yan et al., 2024).

Despite these advancements, LLM-driven translation still exhibits inconsistent quality across languages pairs. While performance often approaches human-level quality for high-resource languages, significant gaps persist for low-resource languages and for linguistically distant or underrepresented pairs (Bawden & Yvon, 2023; Liu et al., 2024; Team et al., 2022; Yan et al., 2024; Zhu et al., 2024). These disparities often result in greater information loss for low-resource languages, underscoring the crucial role of training data. For example, Bang et al. (2023) demonstrate that when translating extremely low-resource languages such as Javanese and Sundanese, ChatGPT often produces literal translations, commits mistranslations or hallucinations, and occasionally defaults to translating into a related but unintended language. Zhu et al. (2024) evaluated and compared popular LLMs across 202 translation directions and found that even the most advanced LLMs lag behind supervised translation models in 83.33% of translation directions—especially in resource-poor scenarios. Moreover, Liu et al. (2024) emphasize that most LLMs remain predominantly English-centric, performing well in English translation but struggling with native-language prompts, particularly in culture-dependent contexts.

These limitations underscore the need for additional training data to improve performance in low-resource settings. However, increasing data availability presents challenges, including the scarcity of native speakers, dialectal diversity, and the absence of standardized writing systems (Aji et al., 2022). A widely adopted approach is to leverage cross-lingual transfer from related languages. Such strategies include using pivot languages, employing transfer learning and engaging in joint training (Neubig & Hu, 2018; Xia et al., 2019). Research has demonstrated that incorporating related languages alongside low-resource languages can enhance translation performance (Neubig & Hu, 2018; Poncelas & Effendi, 2022). This finding underscores the importance of language proximity to evaluate translation quality of LLMs. Recent research has leveraged metrics derived from linguistic databases such as URIEL (Littell et al., 2017), which facilitate more interpretable evaluations through linguistically informed feature vectors that allow for the computation of language distances. These metrics have been applied to select appropriate transfer or pivot languages and to assess language diversity (Lin et al., 2019; Ruder et al., 2021; Nambi et al., 2023).

Language proximity has emerged as a critical factor influencing LLM translation performance. LLMs achieve better cross-lingual transfer between syntactically similar languages than between syntactically distant pairs, as shared structural features facilitate knowledge generalization (Diandaru et al., 2024; Liu et al., 2024). Liu et al. (2024) posited that while LLMs are predominantly English-centric due to imbalanced training corpora, their ability to process non-

English inputs is significantly modulated by the linguistic distance from English. Languages with greater syntactic dissimilarity to English tend to have lower translation performance. In a complementary investigation, Diandaru et al. (2024) explored whether pivoting away from English could enhance translation performance in the Llama 2 model, and the results showed that syntactic similarity is a powerful predictor of translation quality. Notably, even languages with relatively sparse training data, such as Swedish and Catalan, can achieve performance levels comparable to English when linguistic proximity is favorable.

Beyond linguistic proximity, language family also influences translation performance. Translating between languages within the same family, such as English and German, typically yields better results than translating between languages from distinct families (Jiao et al., 2023; Wang et al., 2023). Jiao et al. (2013) demonstrated that ChatGPT's performance declines when translating between unrelated language families, widening the gap between its output and that of commercial translation systems. This challenge is worse for low-resource language pairs; for instance, Wang et al. (2023) reported that translation performance for pairs like Romanian and Chinese suffers significantly due to both resource scarcity and inter-family linguistic differences. These findings suggest that beyond training data availability and linguistic proximity, language family differences present an additional barrier to high-quality translation, particularly in low-resource settings.

Collectively, these observations highlight the need to disentangle the contributions of training data, language proximity, and language family to the multilingual translation capabilities of LLMs. Therefore, the present study aims to examine the interplay among these factors in contributing to information loss during multilingual translation. By systematically evaluating the translation performance of LLMs across language pairs that vary in resource availability and linguistic proximity, our work seeks to address the gap in understanding how these factors collectively affect translation information loss.

To this end, we focus on two state-of-the-art models at the time of testing: GPT-4 and Llama 2. GPT-4, with an estimated 1.76 trillion parameters and trained on approximately 13 trillion tokens from diverse sources, supports a robust array of languages (based on the GPT-3 training dataset) (Abacha et al., 2024). However, detailed information about its training data is not publicly available. Therefore, we also selected an open-source model - Llama 2, available in configurations of 7 billion, 13 billion, and 70 billion parameters, is trained on around 2 trillion tokens and supports a smaller set of languages primarily focused on English and 27 other languages (Touvron et al., 2023).

By investigating the two models, our study aims to explore potential factors influencing multilingual translation qualities in LLMs. In doing so, we seek to understand the mechanisms underlying information loss during multilingual translations in LLMs and provide insights into the limitations of current LLMs. Specifically, we address the following research questions:

1. How does training data interact with language proximity to modulate the translation quality of LLMs?

2. What is the relative importance of specific language distance metrics in predicting translation quality in LLMs?

3. How does language family influence translation quality in LLMs, independent of data resources?

## 2. EXPERIMENT

**2.1 Model Selection**

To conduct the translation tasks, we utilized two large language models (LLMs): GPT-4[1] and Llama 2[2]. While our research references GPT-3's training data, GPT-4 was selected due to its availability at the time of study, as OpenAI had discontinued the API service for GPT-3. This allowed us to investigate the impact of earlier-stage training data on translation performance using the most recent model. Additionally, we selected Llama 2, an open-source model whose training data distribution is publicly available (Touvron et al., 2023). This selection facilitated comparisons between a proprietary model (GPT-4) and an open-source alternative (Llama 2).

---

[1] OpenAI.2023. https://openai.com/index/gpt-4/
[2] Meta. 2023. https://www.llama.com/llama2/

For GPT-4, we selected 88 languages for translation based on GPT-3's training data[3] distribution. Specifically, we chose the 88 languages with the largest word counts in the training data, excluding those contributing less than 0.00005% to the total corpus. For Llama 2, For Llama 2, translations were performed in 26 languages (excluding English), with language selection based on training data where the percentage exceeded 0.05%, as indicated in the language distribution information provided by Touvron et al. (2023)

## 2.2 Prompts

To standardize responses, tailored prompts were designed for each model. For GPT-4, The system prompt was set as:

*Please provide the (language) translation for the following sentence: [Sentence]. Please just show the translation directly without any pronunciation.*

This prompt was adapted from Jiao et al (2023) who demonstrated its effectiveness in optimizing translation outputs. The instruction to exclude pronunciation ensured clarity and focus on the translated text. For Llama 2, the prompt was:

*[Source sentence]*
*Please provide the (language) translation for the above sentence.*
*Please ONLY respond in the following format without any other words:*
*[Original sentence: ORIGINAL_SENTENCE_HERE]*
*[Translation: TRANSLATED_SENTENCE_HERE]*

This design retained the core structure of GPT-4's prompt but introduced explicit formatting instructions to ensure output conciseness.

## 2.3 Materials

The source sentences for translation were drawn from the FLORES-200 dataset (Team et al., 2022). The dataset comprises 3,001 sentences extracted from 852 distinct web articles. These sentences have an average length of approximately 21 words. From this dataset, we randomly selected 1,200 English sentences for a pilot test. After translations were performed using GPT-4 and Llama 2, sentences that Llama 2 could not process due to programming violations or inappropriate content were excluded. For instance: "*I apologize, but I cannot provide a translation for that sentence as it goes against my programming rules to promote or glorify violence or harm towards others.*" This filtering resulted in 1,101 sentences, and we randomly selected 1,000 sentences from them for the final experiment and analysis.

---
[3] https://github.com/openai/gpt-3/blob/master/dataset_statistics/languages_by_word_count.csv

## 2.4 Procedure

To address the limitations of parallel corpora, especially for low-resource languages, we implemented a round-trip translation methodology. This method involves two steps: forward translation (English sentences were translated into target languages) followed by backward translation (the translated sentences were then translated back into English). By performing round-trip translation, we obtained translations both into and from different languages via English, allowing us to compare the original English sentences with the round-trip translations. This approach enables us to evaluate the translation performance of LLMs by assessing the similarity between the source sentence and the round-trip translated sentence. Additionally, this method allows us to analyze the extent of information loss during the process of translating between languages.

The 1,000 English sentences were translated into 88 target languages using GPT-4-0613 via the OpenAI API and into 26 languages using Llama-2-70b-chat-hf via the Hugging Face Inference API. Translations were collected separately with the R package MacBehaviour (Duan et al., 2024). Each sentence was translated independently by the models—one trial per run, with no context or conversation history—repeated five times by both models to account for output variability. The temperature parameter was set to 0.3 for both models to ensure grammatically correct outputs (Ippolito et al., 2019). The maximum token limit for each translation was 500, with all other parameters kept at their default settings. This resulted in 440,000 translations from GPT-4 and 130,000 translations from Llama 2.

Preliminary data cleaning was conducted to address two key issues: redundant information and language mixing. For redundant information, Translations containing additional notes (e.g.,

"Note: Translation may vary depending on context") or instances where the original English sentence was repeated were removed. For language mixing, responses containing elements from multiple languages, particularly in Llama 2's output, were identified and excluded. Three trained student assistants manually reviewed these cases to ensure the final dataset contained only valid translations.

Following data cleaning, the final dataset included 123,702 translated sentences for Llama 2 and 372,136 for GPT-4. These translations were then subjected to backward translation into English. The similarity between the original and back-translated English sentences was assessed using BLEU scores and BERT similarity metrics.

## 2.5 Evaluation

To evaluate translation performance, we employed two metrics: the Bilingual Evaluation Understudy (BLEU) score (Papineni et al., 2002) and Bert similarity. The BLEU score is a widely used measure for assessing the similarity between machine-translated sentences and human-translated reference sentences. It compares the source text and the system output with no reference to their semantic space. We adopted NLTK's default setting with 4-grams using uniform weights for the calculation of the BLUE score. Bert similarity was computed using hidden representations of the embedded sentence pairs, which measures the semantic similarity of the embeddings. For this, we utilized contextual embeddings from a pre-trained encoder-only language model. We used a encoder-only language model dedicated to semantic search as in Sentence-BERT (Reimers and Gurevych, 2019) for the calculation of Bert similarity, as its Siamese network architecture enables fixed-sized vectors as the input, mitigating existing issues when using the averaged BERT layers or the representation of the [CLS] token. Following the approaches above, evaluations were performed by comparing a back-translated sentence (in English) and its source sentence (in English).

We also measured the distance between English and each of the target languages. Language distance, as in He et al. (2019), was quantified using a range of metrics from the URIEL knowledge base (Littell et al., 2017), including phylogenetic, geographic, syntactic, phonological, featural, and (phonetic) inventory distances. The syntactic, phonological, and inventory features are derived from existing language documentation. These binary features denote the presence or absence of certain syntactic properties, phonology properties, or phonemes, respectively. The phylogenetic distance is based on language families, and the geographic distance represents the distance of geographical location of languages. Additionally, orthographic distance was quantified by measuring the Levenshtein distance between English words and their translation equivalents in Swadesh lists, which comprise 207 core vocabulary items from each target language (available from the Natural Language Toolkit; Bird et al., 2009). Languages which are more closely related to English tend to spell translation equivalents more similarly (e.g., the German word for "nose" is "Nase"). Longer words tend to have larger distances, so to normalize for length, orthographic distance was divided by the maximum possible distance, for a value between 0 (identical forms) and 1 (forms that differ in all letters), and then averaged across the 207 Swadesh list items (Kumar et al. 2022). Along with measuring phylogenetic distance from English, orthographic distance accounts for whether a language uses the Latin alphabet, since languages that use other scripts have the maximum normalized distance of 1.

## 2.6 Results

To evaluate the influence of training data and language distance on translation quality, we performed log transformation followed by scaling for both training data and language distance. For translations generated by LLaMA 2, the BLEU score was highest when English was translated into and then back from Indo-European-Germanic languages ($M = 0.48$, $SD = 0.21$), followed by Indo-European-Romance languages ($M = 0.48$, $SD = 0.21$), Indo-European-Slavic languages ($M = 0.39$, $SD = 0.19$), Austronesian languages ($M = 0.35$, $SD = 0.18$), Other languages ($M = 0.28$, $SD = 0.16$), and Sino-Tibetan languages ($M = 0.25$, $SD = 0.14$) (see also Figure 1).

For BERT similarity, Indo-European-Romance languages scored the highest ($M = 0.93$, $SD = 0.09$), followed by Indo-European-Germanic languages ($M = 0.93$, $SD = 0.09$), Indo-European-Slavic languages ($M = 0.90$, $SD = 0.12$), Austronesian languages ($M = 0.89$, $SD = 0.11$), Other languages ($M = 0.85$, $SD = 0.13$), and Sino-Tibetan languages ($M = 0.84$, $SD = 0.13$) (see Figure 2).

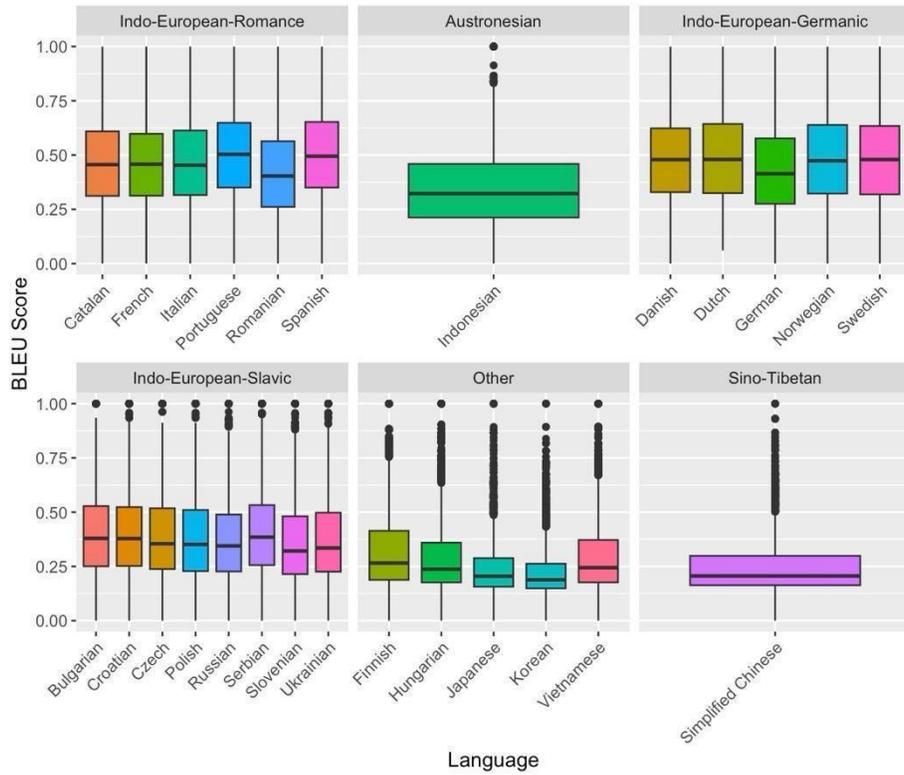

**Figure 1**. BLEU score of Llama2 summarized by language and family

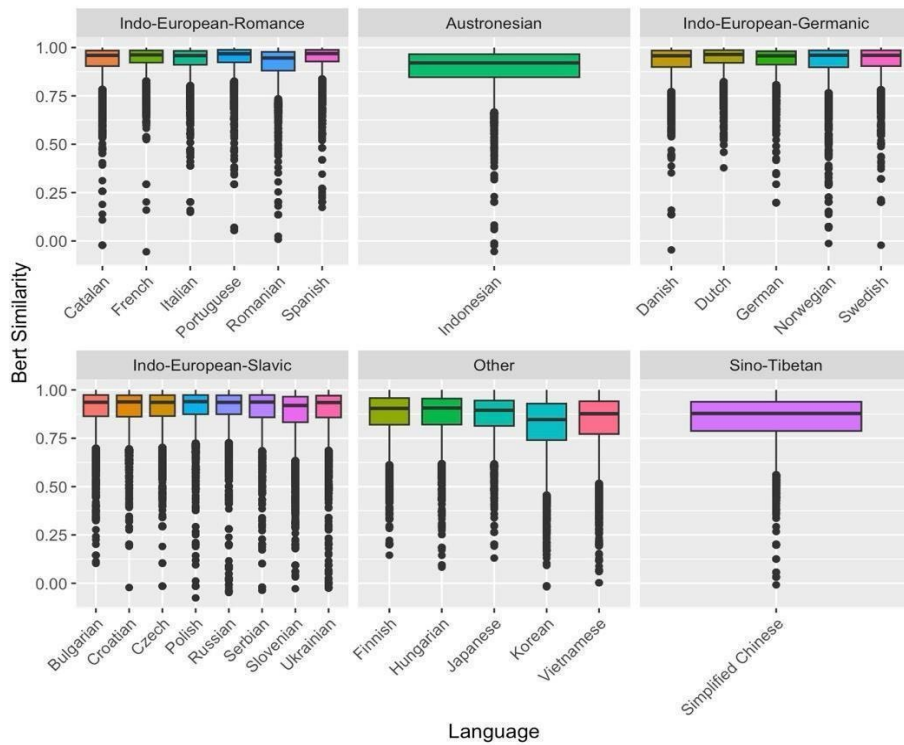

**Figure 2**. Bert similarity of Llama2 summarized by language and family

For translations produced by GPT-4, the BLEU score was highest for Indo-European-Romance languages ($M = 0.54$, $SD = 0.20$), followed by Indo-European-Germanic ($M = 0.48$, $SD = 0.22$), Indo-European-Slavic ($M = 0.44$, $SD = 0.20$), Austronesian ($M = 0.33$, $SD = 0.19$), Indo-European-Other ($M = 0.32$, $SD=0.20$), Afro-Asiatic ($M = 0.37$, $SD = 0.18$), Other ($M = 0.28$, $SD = 0.17$), Sino-Tibetan ($M = 0.24$, $SD = 0.15$), Turkic ($M = 0.25$, $SD = 0.15$), Indo-European-Indo-Aryan ($M = 0.23$, $SD = 0.14$), Dravidian ($M = 0.19$, $SD = 0.09$), Atlantic-Congo ($M = 0.27$, $SD = 0.18$), and Niger-Congo languages ($M = 0.16$, $SD = 0.06$) (see Figure 3).

For BERT similarity, Indo-European-Romance languages scored the highest ($M = 0.96$, $SD = 0.06$), followed by Indo-European-Slavic ($M = 0.93$, $SD = 0.09$) and Indo-European-Germanic ($M = 0.89$, $SD = 0.17$), Afro-Asiatic ($M = 0.84$, $SD = 0.17$), Austronesian ($M = 0.77$, $SD = 0.25$), Indo-European-Other ($M = 0.75$, $SD = 0.26$), Turkic ($M = 0.71$, $SD = 0.25$), Other ($M = 0.70$, $SD = 0.31$), Dravidian ($M = 0.65$, $SD = 0.22$), Indo-European-Indo-Aryan ($M = 0.66$, $SD = 0.30$), Sino-Tibetan ($M = 0.60$, $SD = 0.37$), Atlantic-Congo ($M = 0.57$, $SD = 0.36$), and Niger-Congo languages ($M = 0.36$, $SD = 0.25$) (see Figure 4).

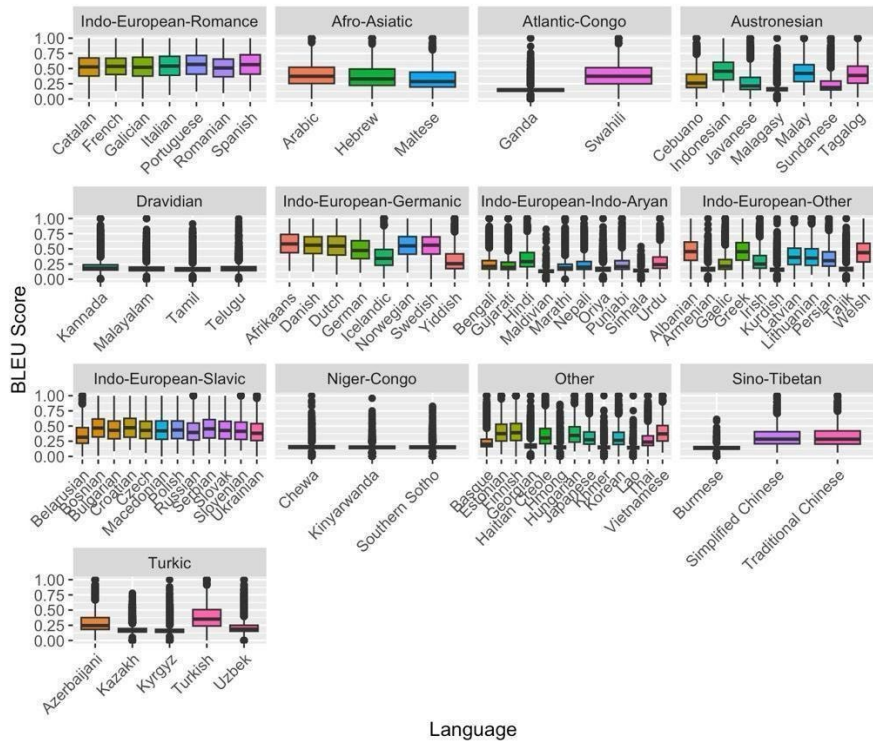

**Figure 3.** BLEU score of GPT-4 summarized by language and family

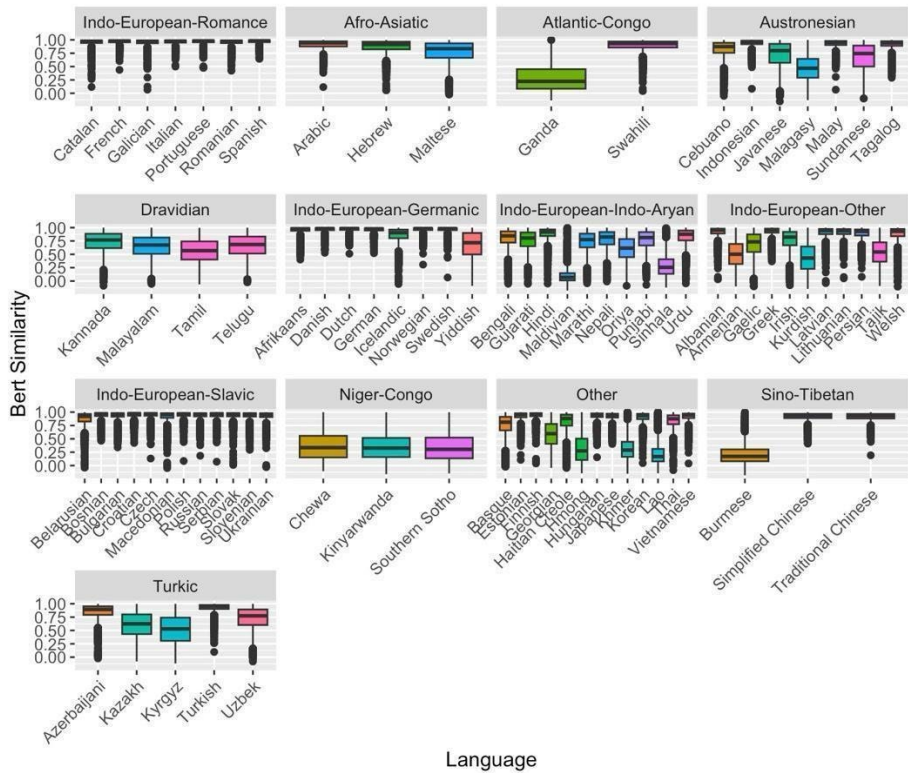

**Figure 4**. BERT similarity of GPT-4 summarized by language and family

We used linear mixed-effects regression models implemented in the lme4 package (Bates et al., 2015) in R (R Core Team, 2014) to investigate the impact of training data and language distance on translation quality, as measured by BLEU score and BERT similarity. Fixed effects included training data and average language distance (calculated from phylogenetic, geographical, syntactic, phonological, featural, inventory, and orthographic distances), with translation sentences as random effects. Significance was assessed using the lmerTest package (Kuznetsova et al., 2017). We used a forward algorithm to determine the maximal random effect structure based on the data, using the alpha level of 0.2 instead of 0.05 (Matuschek et al., 2017). Collinearity checks were performed prior to analysis.

For BLEU scores of LLaMA 2 translations (Figure 5), significant main effects were observed for language distance ($\beta = -0.05$, $SE = 0.001$, $t = -39.60$, $df = 1{,}091$, $p < 0.001$) and training data ($\beta = 0.02$, $SE = 0.001$, $t = 46.04$, $df = 121{,}805$, $p < 0.001$). A significant interaction between language distance and training data was also found ($\beta = 0.02$, $SE = 0.0005$, $t = 39.82$, $df = 121{,}905$, $p < 0.001$). The same pattern was observed for BERT similarity (Figure 6). We found significant main effects of language distance ($\beta$ = -0.017, $SE$ = 0.0006, $t$ = -28.26, $df$ = 1111, $p$ < 0.001) and training data ($\beta$ = 0.013, $SE$ = 0.0003, $t$ = 45.01, $df$ = 121805, $p$ < 0.001). A significant interaction between language distance and training data was also found ($\beta$ = 0.007, $SE$ = 0.0003, $t$ = 26.61, $df$ = 121905, $p$ < 0.001).

For GPT-4 BLEU scores (Figure 7), language distance ($\beta$ = -0.048, $SE$ = 0.0003, $t$ = -140.30, $df$ = 370,705, $p$ < 0.001) and training data ($\beta$ = 0.071, $SE$ = 0.0003, $t$ = 207.24, $df$ = 371,205, $p$ < 0.001) showed significant main effects, with a significant interaction ($\beta$ = 0.006, $SE$ = 0.0006, $t$ = 10.57, $df$ = 943, $p$ < 0.001). The same pattern was observed for BERT similarity (Figure 8), significant main effects were found for language distance ($\beta$ = -0.067, $SE$ = 0.0005, $t$ = -146.91, $df$ = 370,305, $p$ < 0.001) and training data ($\beta$ = 0.135, $SE$ = 0.0014, $t$ = 96.12, $df$ = 1,084, $p$ < 0.001), with a significant interaction ($\beta$ = 0.055, $SE$ = 0.0003, $t$ = 177.34, $df$ = 337,005, $p$ < 0.001).

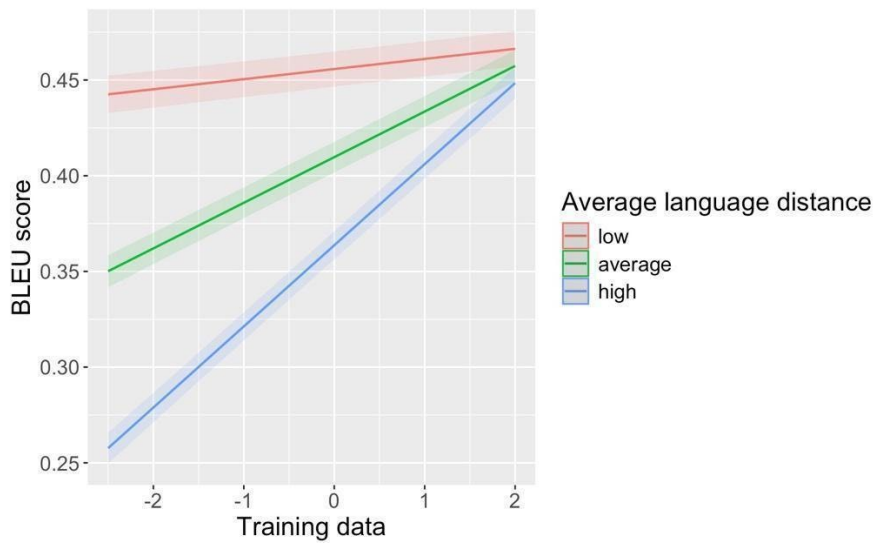

**Figure 5.** The relation between training data and BLEU score grouped by language distances (Llama2)

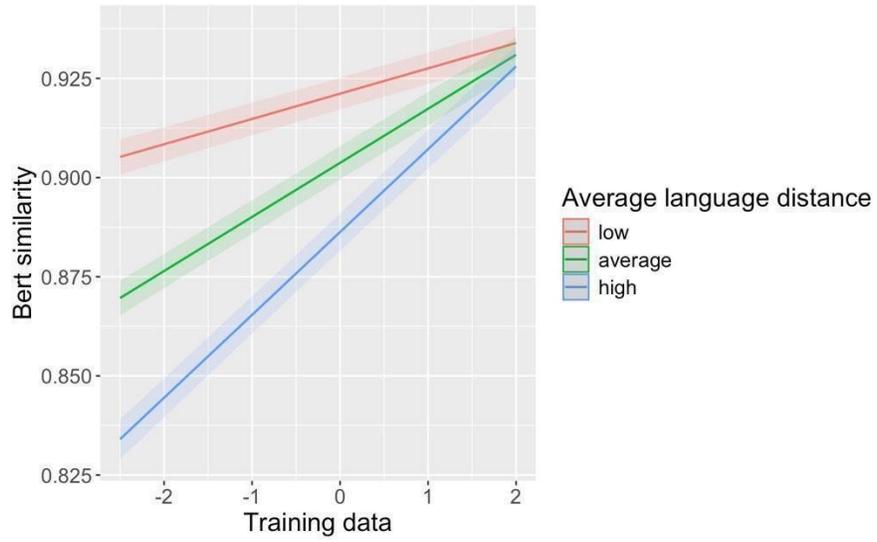

**Figure 6.** The relation between training data and Bert similarity grouped by language distances (Llama2)

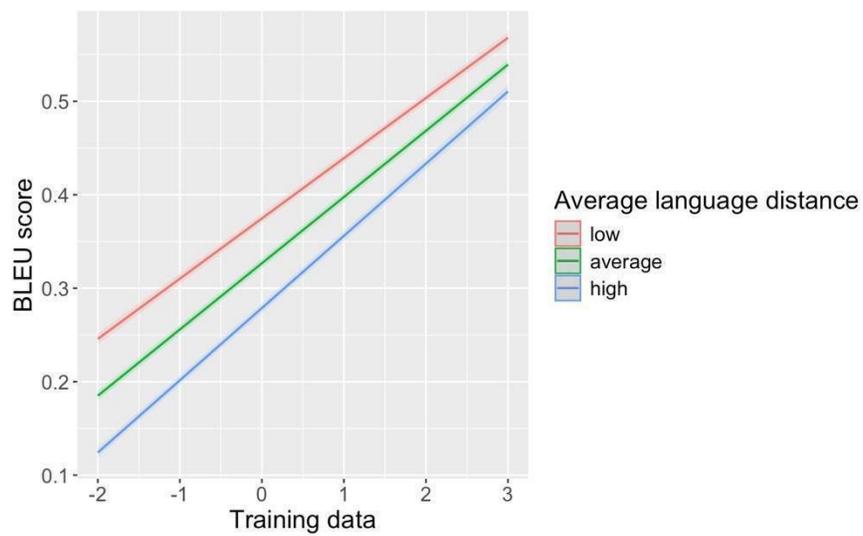

**Figure 7.** The relation between training data and BLEU score grouped by language distances (GPT-4)

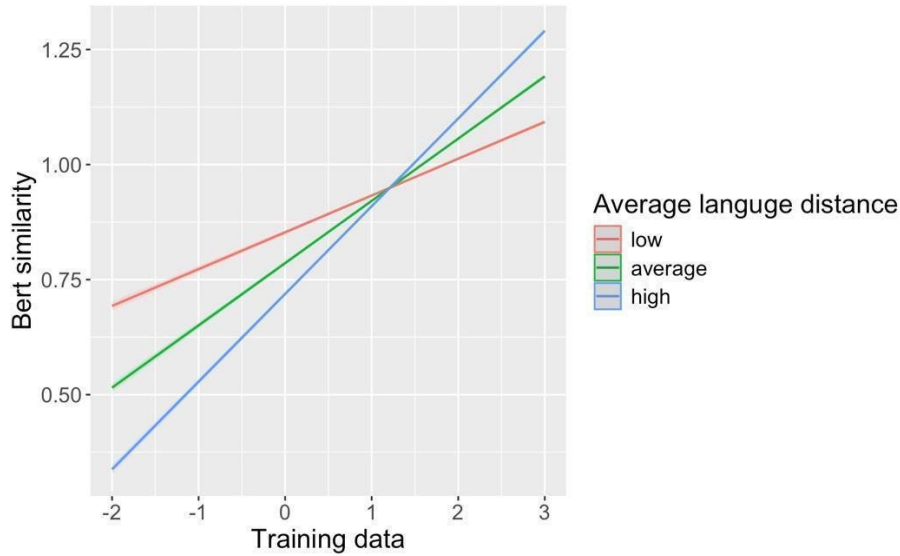

**Figure 8.** The relation between training data and Bert similarity grouped by language distances (GPT-4)

To further evaluate the relative importance of specific language distance metrics in predicting translation quality, we conducted random forest analyses (Figure 9). For LLaMA 2, phylogenetic distance and orthographic distance emerged as the most important predictors of BLEU scores, while geographical and syntactic distances were moderately influential. The predictors featural distance, phonological distance and inventory distance contributed less to the model. For LLaMA 2's BERT similarity, orthographic distance was identified as the strongest predictor, followed by phylogenetic distance and geographical distance. Syntactic distance, featural distance, phonological distance, and inventory distance exhibited relatively lower levels of predictive influence.

For GPT's BLEU scores, orthographic distance and geographical distance emerged as the most significant predictors. Syntactic distance also played a prominent role, while phylogenetic distance, featural distance, inventory distance, and phonological distance demonstrated lower levels of importance. For GPT's BERT similarity, geographical distance was the most impactful predictor, followed by syntactic distance and orthographic distance. Featural distance, inventory distance, phylogenetic distance, and phonological distance demonstrated comparatively lower contributions to the model.

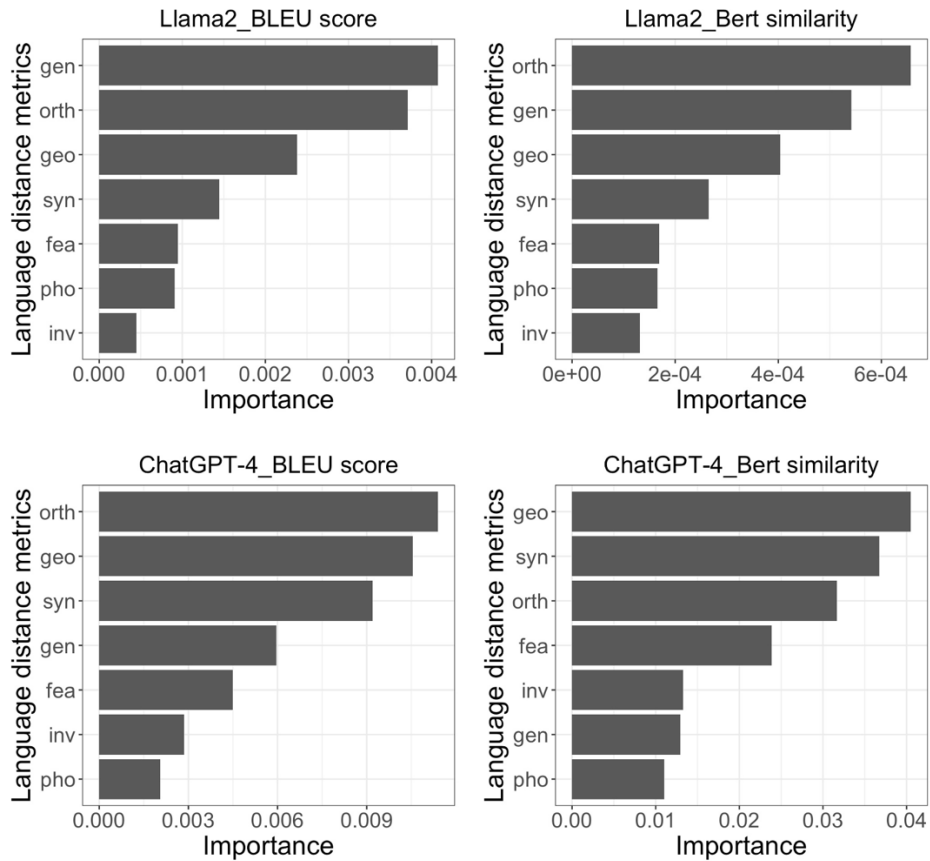

**Figure 9.** Importance of specific language distance metrics in explaining translation quality of Llama2 and GPT-4

To examine the effect of language family on translation quality independent of training data, we conducted a mixed-effects model analysis using the residuals of a prior model as the dependent variable. The prior model included training data as the independent variable and BLEU score as the dependent variable. In the current analysis, language family was included as a fixed effect, while 'translation sentences' was treated as a random effect. Indo-European-Romance was set as the reference level for the fixed effect.

As shown in Figure 10, for LLaMA 2, BLEU scores of the Indo-European-Romance family were significantly lower than those of the Indo-European-Germanic family ($\beta = 0.01$, $SE = 0.0012$, $t = 10.19$, $p < 0.001$), while they were significantly higher than those of the Austronesian ($\beta = -0.08$, $SE = 0.0022$, $t = -36.33$, $p < 0.001$), Indo-European-Slavic ($\beta = -0.03$, $SE = 0.0011$, $t = -$

25.03, $p < 0.001$), Other ($\beta = -0.16$, $SE = 0.0013$, $t = -123.81$, $p < 0.001$), and Sino-Tibetan families ($\beta = -0.16$, $SE = 0.0023$, $t = -68.16$, $p < 0.001$).

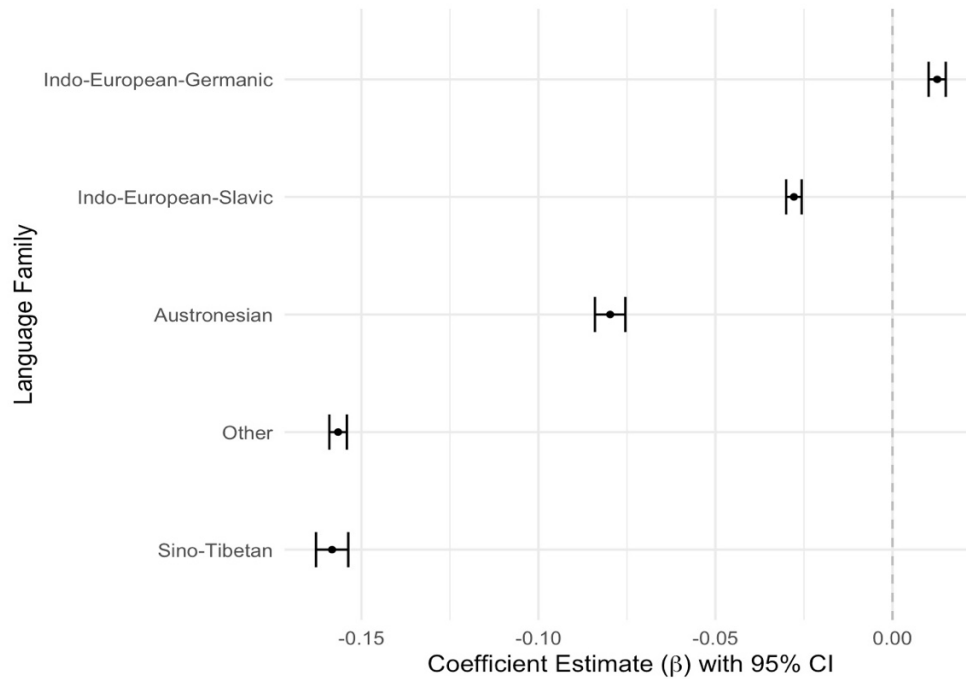

**Figure 10.** Forest plot of language family effects on translation quality residuals (BLEU score of Llama2)

As shown in Figure 11, for LLaMA 2, Bert similarity of the Indo-European-Romance family did not significantly differ from that of the Indo-European-Germanic family ($\beta = 0.0013$, $SE = 0.00075$, $t = 1.78$, $p = 0.07$), but was significantly higher than that of the Austronesian ($\beta = -0.020$, $SE = 0.0013$, $t = -15.18$, $p < 0.001$), Indo-European-Slavic ($\beta = -0.0062$, $SE = 0.00067$, $t = -9.29$, $p < 0.001$), Other ($\beta = -0.058$, $SE = 0.00076$, $t = -76.19$, $p < 0.001$), and Sino-Tibetan families ($\beta = -0.053$, $SE = 0.0014$, $t = -37.60$, $p < 0.001$).

As shown in Figure 12, for GPT-4, BLEU scores of the Indo-European-Romance family were significantly lower than those of the Indo-European-Germanic family ($\beta = 0.012$, $SE = 0.0013$, $t = 8.98$, $p < 0.001$), while they were significantly higher than those of the Afro-Asiatic ($\beta = -0.033$, $SE = 0.0016$, $t = -20.46$, $p < 0.001$), Atlantic-Congo ($\beta = -0.018$, $SE = 0.0018$, $t = -10.03$, $p < 0.001$), Austronesian ($\beta = -0.048$, $SE = 0.0013$, $t = -37.84$, $p < 0.001$), Dravidian ($\beta = -0.17$, $SE = 0.0014$, $t = -118.96$, $p < 0.001$), Indo-European-Indo-Aryan ($\beta = -0.088$, $SE = 0.0012$, $t = -74.48$, $p < 0.001$), Indo-European-Other ($\beta = -0.052$, $SE = 0.0012$, $t = -43.82$, $p < 0.001$),

Niger-Congo ($\beta = -0.14$, $SE = 0.0016$, $t = -89.83$, $p < 0.001$), Other ($\beta = -0.12$, $SE = 0.0012$, $t = -100.29$, $p < 0.001$), Sino-Tibetan ($\beta = -0.17$, $SE = 0.0017$, $t = -101.46$, $p < 0.001$), and Turkic families ($\beta = -0.10$, $SE = 0.0014$, $t = -70.56$, $p < 0.001$). BLEU scores of the Indo-European-Romance family did not significantly differ from those of the Indo-European-Slavic family ($\beta = -0.0015$, $SE = 0.0012$, $t = -1.26$, $p = 0.21$).

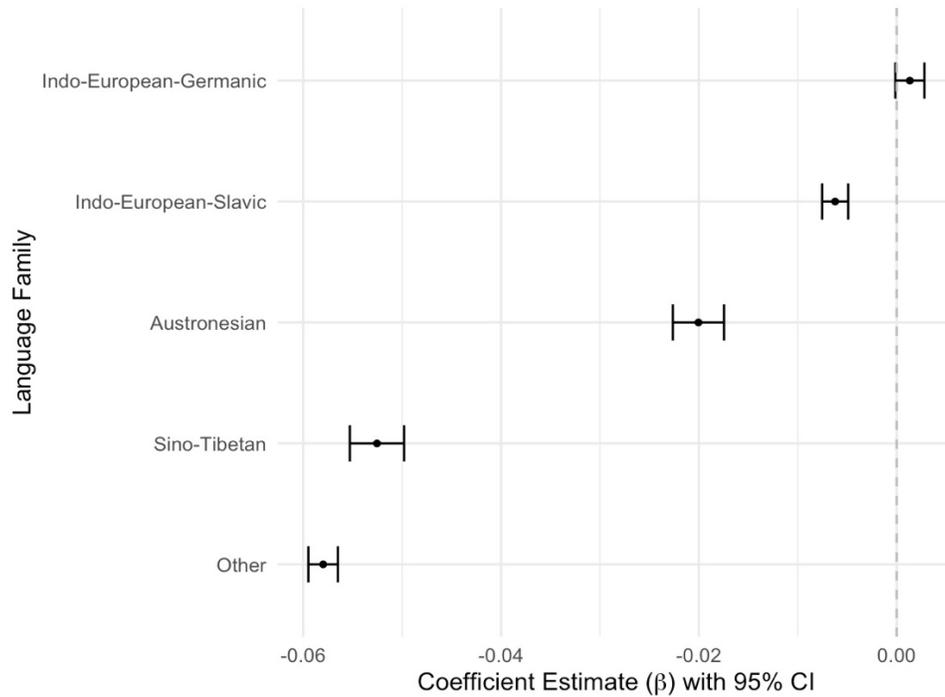

**Figure 11.** Forest plot of language family effects on translation quality residuals (BERT similarity of Llama2)

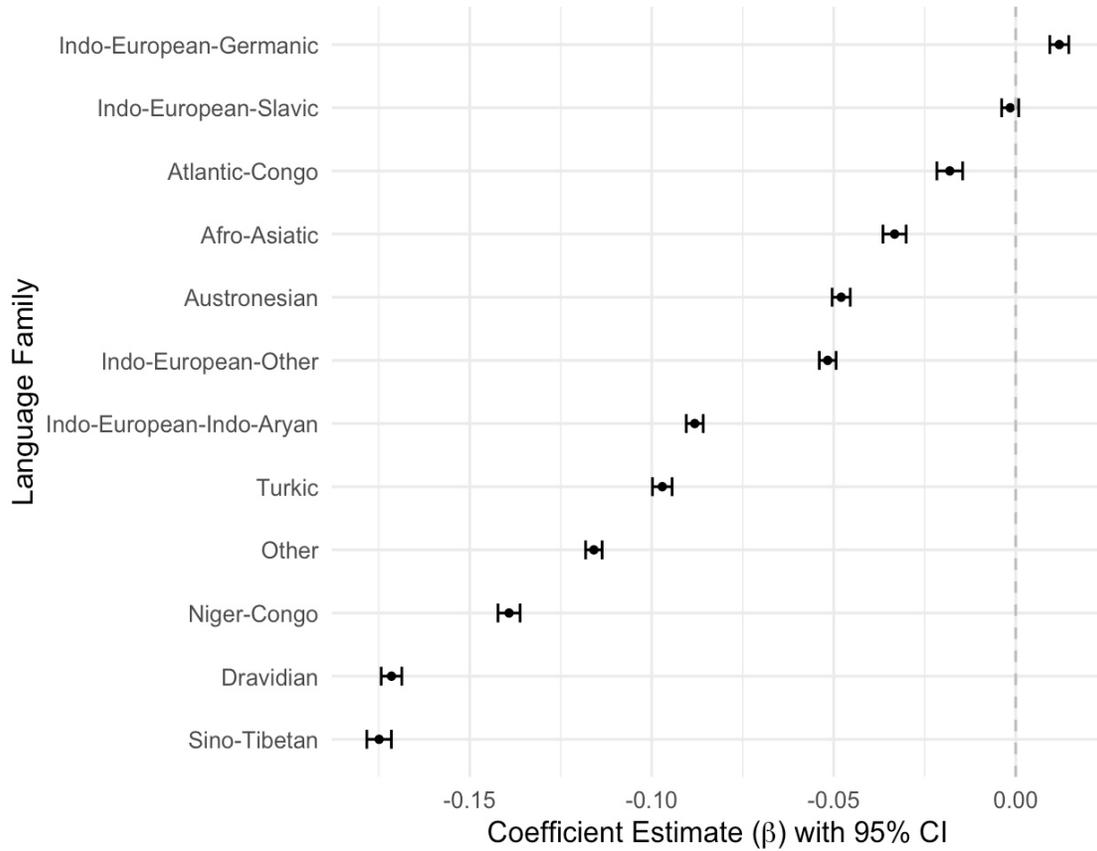

**Figure 12.** Forest plot of language family effects on translation quality residuals (BLEU score of GPT-4)

As shown in Figure 13, for GPT-4, Bert similarity of the Indo-European-Romance family was significantly lower than that of the Afro-Asiatic ($\beta = 0.11$, $SE = 0.0023$, $t = 48.21$, $p < 0.001$), Atlantic-Congo ($\beta = 0.016$, $SE = 0.0025$, $t = 6.38$, $p < 0.001$), Austronesian ($\beta = 0.082$, $SE = 0.0018$, $t = 46.00$, $p < 0.001$), Indo-European-Germanic ($\beta = 0.045$, $SE = 0.0019$, $t = 23.89$, $p < 0.001$), Indo-European-Indo-Aryan ($\beta = 0.049$, $SE = 0.0017$, $t = 29.62$, $p < 0.001$), Indo-European-Other ($\beta = 0.059$, $SE = 0.0017$, $t = 35.28$, $p < 0.001$), Indo-European-Slavic ($\beta = 0.13$, $SE = 0.0017$, $t = 75.20$, $p < 0.001$), and Turkic families ($\beta = 0.07$, $SE = 0.0019$, $t = 35.93$, $p < 0.001$), while it was significantly higher than that of the Dravidian ($\beta = -0.03$, $SE = 0.0020$, $t = -13.01$, $p < 0.001$), Niger-Congo ($\beta = -0.21$, $SE = 0.0022$, $t = -98.21$, $p < 0.001$), Other ($\beta = -0.03$, $SE = 0.0016$, $t = -16.61$, $p < 0.001$), and Sino-Tibetan families ($\beta = -0.16$, $SE = 0.0024$, $t = -67.32$, $p < 0.001$).

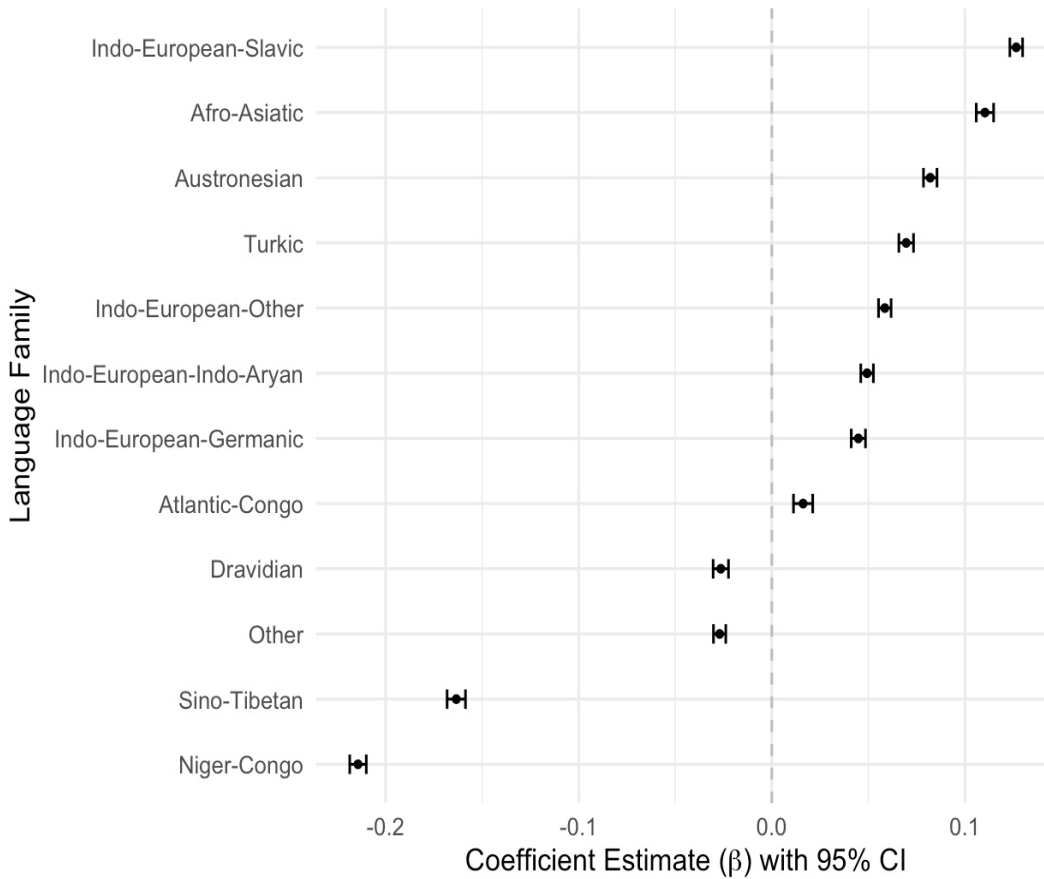

**Figure 13.** Forest plot of language family effects on translation quality residuals (BERT similarity of GPT-4)

## 3 DISCUSSION

We systematically examined the factors contributing to information loss during multilingual translation in LLMs. Our results revealed how training data interact specifically with language proximity, the relative importance of distinct language distance metrics, and the independent impact of language family on multilingual translation quality. Specifically, we found that: (1) significant interactions exist between training data and language distance for both GPT-4 and LLaMA 2, demonstrating that abundant training data can partially compensate for the disadvantages posed by linguistic differences, while linguistic proximity to English mitigates information loss in low-resource scenarios; (2) although the relative importance of specific language distance metrics in predicting translation quality of LLMs varied slightly between language models and quality metrics, overall, orthographic, phylogenetic, geographical, and syntactic distances contributed more to predicting translation quality, with phonetic, inventory and featural distances less influential; and (3) language family independently influenced LLMs' translation quality, with Indo-European-Romance and Indo-European-Germanic languages generally exhibiting higher translation quality compared to linguistically distant families.

**3.1 Interaction of Training Data and Language Proximity**

We found significant interactions between training data and language distance from English, influencing translation quality for both GPT-4 and LLaMA 2. Specifically, in low-resource conditions, translation quality consistently improved as language proximity to English increased. However, with sufficient training data, linguistically distant languages did not necessarily show poorer translation quality; in some cases, they demonstrated comparable or even superior translation quality relative to closer languages. This suggests the compensatory role that abundant training data can play in overcoming linguistic barriers. Increased linguistic exposure enables models to better understand and generate language-specific patterns, even for linguistically distant languages. Additionally, compared with LLaMA 2, GPT-4, with its larger model size and broader training corpus diversity, demonstrated a stronger ability to maintain translation quality across linguistically diverse languages.

Our findings also highlight the advantage of linguistic proximity, particularly under conditions of limited training data. The observation that languages closer to English yield higher translation accuracy when resources are scarce suggests that similarity facilitates knowledge transfer due to shared linguistic characteristics (Poncelas & Effendi, 2022). Structural and lexical similarities among languages significantly ease the burden of learning for translation models by providing more direct analogies and transferable linguistic patterns. Linguistic proximity implies shared vocabulary roots, similar grammatical structures, and comparable sentence construction patterns, which collectively minimize the cognitive and computational load on models by reducing the complexity of cross-lingual mappings. When linguistic resources are limited, these similarities act as critical scaffolds, enabling models to leverage their existing linguistic knowledge more efficiently. For instance, even languages with limited resources, such as Swedish and Catalan, could achieve comparable translation quality to English due to their syntactic proximity (Diandaru et al., 2024), emphasizing the critical advantage of linguistic similarity in translation tasks, particularly under constrained data conditions. These findings also aligned with Liu et al. (2024) who observed a significant correlation between syntactic distance and translation performance, indicating the crucial role of syntactic similarities in supporting effective cross-lingual transfer.

### 3.2 Importance of Specific Language Distance Metrics

Language distance metrics demonstrated different importance in predicting multilingual translation quality within LLMs. Orthographic, phylogenetic, geographical, and syntactic distances emerged as highly significant predictors, whereas phonetic, inventory, and featural distances exhibited relatively minor influences.

Orthographic distance was particularly predictive across models, possibly due to its direct interaction with core architectural mechanisms such as subword tokenization. The use of subword tokenization is instrumental in handling diverse vocabularies and linguistic structures. Subword tokenization divides text into smaller, meaningful units, effectively managing rare or out-of-vocabulary words (Haslett & Cai, 2024). This approach is advantageous for languages with shared roots or similar orthographic patterns, as it enables the model to build translations using familiar subword components. However, for morphologically rich or orthographically distant languages, subword tokenization can introduce challenges. Languages with complex inflections or scripts vastly different from English may produce fragmented tokens, where single words are split into numerous subword units. This fragmentation increases the computational burden on the model and can dilute the semantic coherence of the translation. Consequently, languages with high orthographic divergence tend to have lower BLEU and BERT similarity scores, as tokenization issues propagate through the translation process. For example, in languages that do not use the Latin script, such as Sino-Tibetan or Niger-Congo families, subword tokenization often struggles to capture meaningful linguistic units, resulting in reduced translation fidelity. These challenges highlight the importance of optimizing tokenization strategies for diverse linguistic structures, particularly in multilingual models. Future advancements in adaptive tokenization mechanisms tailored to specific language families may mitigate these effects.

Phylogenetic distance, reflecting historical linguistic relatedness and common ancestry, also emerged as a crucial predictor of translation quality. Phylogenetically similar languages possess similarities in lexical items, morphological structures, and syntactic frameworks. These similarities may enhance the capacity of LLMs to perform effective cross-linguistic knowledge transfer. This is supported by Mekki and Abdul-Mageed (2024) who reported that phylogenetic distance strongly correlates with translation performance, particularly in low-resource language settings, indicating that phylogenetically closer languages are generally translated more accurately by LLMs. Stap et al. (2023) further demonstrated that both phylogenetic and syntactic distance are among the most significant linguistic distance predictors influencing knowledge transfer in multilingual machine translation. These findings collectively highlight phylogenetic distance as a critical factor for translation quality within multilingual contexts.

Geographic distance also plays a critical role in shaping cross-linguistic similarities relevant to multilingual translation. Similarly, language diversity in general increases as geographic distance increases (Huisman, Majid & Van Hout, 2019). The use of language in human interaction can be thought of as linguistic gene flow. This interaction will be more intense between people that are close to each other, leading to extensive borrowing of lexical items, convergence of syntactic structures, and alignment in semantic and pragmatic usage patterns. As a result, the language of neighboring communities will differ only slightly (Chambers and Trudgill, 1998), inherently sharing linguistic features that LLMs can exploit effectively, thereby enhancing the accuracy and coherence of multilingual translation. In contrast, geographically distant communities experience less frequent interaction and limited linguistic accommodation. This reduction in linguistic gene flow over increasing distances results in less structural and lexical resemblance among speech communities, which poses greater challenges for LLMs in aligning representations across languages.

Consistent with our own findings, which showed syntactic distance as one of the most predictive linguistic features influencing translation quality, existing research has similarly emphasized its importance. For instance, Diandaru et al. (2024) also identified syntactic distance as one of the most significant predictors of translation accuracy in multilingual tasks involving LLMs, and Dryer & Haspelmath (2013) found that translation accuracy significantly correlates with syntactic distance but not with other types of distances. These are further demonstrated by

Stap et al. (2023), who found that syntactic distance strongly influences the effectiveness of knowledge transfer in multilingual translation systems. The underlying mechanism behind this relationship probably resides in how LLMs generalize linguistic structures across languages. Specifically, syntactic similarity facilitates more efficient cross-lingual transfer by enabling the activation of overlapping internal representations. Since LLMs predict tokens based on contextual patterns learned during pretraining, structural similarities across languages allow models to map syntactic templates more effectively. Conversely, Variations in syntactic structures, such as differences in word order poses greater challenges. These syntactic divergences require LLMs to perform extensive structural transformations, substantially increasing cognitive and computational demands and potentially introducing translation errors and information loss.

It is noteworthy that syntactic distance emerged as relatively more influential in predicting translation quality in GPT-4 compared to LLaMA 2. This heightened sensitivity in GPT-4 could be attributed to its larger scale and more sophisticated architecture, which may better capture intricate syntactic features during extensive pre-training. GPT-4's advanced self-attention mechanisms potentially allow it to implicitly model complex syntactic structures more effectively, thereby making syntactic variations more influential in its translation performance. Conversely, phylogenetic distance appeared more influential for LLaMA 2, likely reflecting its relatively smaller model size and potentially narrower exposure to diverse linguistic data. Smaller-scale models might rely more heavily on lexical and morphological similarities provided by phylogenetic relatedness to compensate for their comparatively limited capacity to model deeper syntactic structures.

Phonetic, inventory, and featural distances exhibited relatively minor influences on the translation quality of large language models, primarily due to their linguistic properties and the nature of LLM translation tasks. These distances predominantly capture variations at the speech and sound level, including differences in phoneme sets and pronunciation patterns. Since LLMs fundamentally operate on textual data rather than spoken language, phonetic and phonological differences have limited direct relevance for text-based translation tasks.

**3.3 Independent Influence of Language Family**

Beyond individual linguistic metrics, language family itself had a distinct, independent influence on translation quality, indicating that intrinsic linguistic features of specific language families influence how effectively large language models generalize learned linguistic knowledge.

The mechanisms behind LLMs may provide explanations to the influence of language family. The self-attention mechanism, a fundamental unit of transformer-based architecture, enables the model to capture long-range dependencies within and across sentences. This is particularly beneficial for translation tasks, where maintaining coherence and accurately reflecting syntactic relationships are critical. Self-attention allows the model to weigh the importance of different words in a sentence, facilitating context-sensitive translation. languages within the same family typically share common lexical roots, morphological structures, and syntactic frameworks (Campbell and Poser, 2008), and they are better positioned to benefit from this mechanism, which collectively facilitates the generalization capabilities of LLMs. For example, languages belonging to the Indo-European-Romance and Indo-European-Germanic families display substantial lexical and structural overlaps with English, including shared vocabulary derived from Latin or common Germanic roots, and similarities in grammatical constructs, enabling efficient mapping through specialized attention heads. These heads, often tuned to English or related languages during training, enhance the model's ability to process similar languages by leveraging learned patterns and relationships. This specification, referred to as language-specialized attention heads, might extend to closely related languages, streamlining their processing (Zeng et al., 2024; Zhang et al., 2024). Sentences shared identical grammatical rules (with different semantic meanings) showed similar activation patterns in hidden states among LLMs (Zhou et al., 2024), which further confirmed the specialization in transformer architecture. However, for languages with different families, the lack of shared specialization with English may increase the complexity and cognitive load for models during translation tasks and hinder effective generalization, reducing translation quality.

Interestingly, although Indo-European Romance and Germanic are the language families structurally and lexically closest to English, GPT-4 did not achieve the highest BERT similarity scores on these families. This outcome likely reflects the model's adaptive generation strategies. Specifically, GPT-4 may generate translations that are semantically faithful but structurally divergent from the reference, particularly when handling languages from families that are closely related to English. These translations often involve paraphrasing, syntactic restructuring, or stylistic variation—features that, while linguistically appropriate, reduce surface similarity with reference translations and thereby lower BERT-based semantic scores. In contrast, when translating into languages from more distant families, the model may adopt more literal strategies, producing outputs that are closer in form to the reference and thus scoring higher on surface-sensitive metrics like BERT similarity.

Our findings indicate that as model capacity increases, translation behavior becomes more flexible and context-sensitive—but such flexibility may not always align with evaluation metrics. More importantly, this finding suggests that advanced LLMs such as GPT-4 exhibit language-family-sensitive adaptation in translation strategy. These models modulate their behavior according to the structural characteristics of the target language family, becoming more form-preserving for distant families and more semantically expressive for closely related ones. This language-family-sensitive behavior indicates that LLMs may exhibit early forms of language awareness in their generative processes.

## 4 CONCLUSION

This study systematically examined the factors contributing to information loss in multilingual translation by LLMs, focusing on the interaction between training data, language proximity, and language family. By evaluating GPT-4 and LLaMA 2 across typologically diverse languages, we demonstrated that translation quality is jointly modulated by data availability and linguistic proximity. While abundant training data can compensate for linguistic dissimilarities, languages structurally closer to English tend to yield better performance in low-resource scenarios. Moreover, language distance metrics, particularly orthographic, phylogenetic, syntactic, and geographical distance, emerged as more informative predictors of translation quality. Notably, language family exerted an independent influence on model performance even when controlling training data. Although Indo-European-Romance and Germanic languages typically scored higher, GPT-4 showed lower BERT similarity on these families, suggesting that advanced LLMs such as GPT-4 exhibit language-family-sensitive adaptation in translation strategy.

Practically, these findings offer insights into multilingual model development. In low-resource contexts, translation performance could be enhanced by leveraging linguistically similar languages as transfer languages, while linguistically distant languages may benefit more from tailored fine-tuning or targeted data augmentation strategies. Future efforts may explore the integration of syntactic representations into model architectures and how LLMs tailor their translation strategies to different language types.

Taken together, these findings contribute to a deeper understanding of the linguistic constraints shaping multilingual translation in LLMs. They emphasize that translation quality is not merely a function of data volume, but is deeply rooted in structural, typological, and representational relationships between languages. Moving forward, more inclusive and linguistically informed approaches to model training, evaluation, and design may be beneficial for advancing the multilingual translation capabilities of large language models.